\documentclass[sigconf]{acmart}
\usepackage{siunitx}
\usepackage[utf8]{inputenc}
\usepackage[T1]{fontenc}
\usepackage{newunicodechar}
\usepackage[table]{xcolor}
\usepackage{multirow}
\usepackage{amsmath}
\newunicodechar{、}{,}
\AtBeginDocument{%
  }

\setcopyright{acmlicensed}
\copyrightyear{2026}
\acmYear{2026}
\acmDOI{XXXXXXX.XXXXXXX} 
\acmConference[ACM MM 2026]{The 34th ACM International Conference on Multimedia}{November 10--14, 2026}{Rio de Janeiro, Brazil}
\acmISBN{978-1-4503-XXXX-X/2026/11} 

\begin{document}

\title{ALAS: Adaptive Long-Horizon Action Synthesis via Async-pathway Stream Disentanglement}

\author{Yutong Shen}
\affiliation{%
  \institution{Beijing University of Technology}
  \city{Beijing}
  \country{China}
}
\email{syt2004@emails.bjut.edu.cn}

\author{Hangxu Liu}
\affiliation{%
  \institution{Fudan University}
  \city{Shanghai}
  \country{China}}
\email{23210720103@m.fudan.edu.cn}

\author{Lei Zhang}
\affiliation{%
  \institution{ University of Hamburg}
  \city{Hamburg}
  \country{Germany}
}
\email{liupenghui@emails.bjut.edu.cn}

\author{Penghui Liu}
\affiliation{%
  \institution{Beijing University of Technology}
  \city{Beijing}
  \country{China}
}
\email{liupenghui@emails.bjut.edu.cn}

\author{Yinqi Liu}
\affiliation{%
  \institution{Beijing University of Technology}
  \city{Beijing}
  \country{China}
}
\email{lllllyq3101@qq.com}

\author{Liuxiang Yang}
\affiliation{%
  \institution{Hubei University of Chinese Medicine}
  \city{Wuhan}
  \country{Hubei}
}
\email{2023307013752@stmail.hbucm.edu.cn}

\author{Tongtong Feng}
\affiliation{%
 \institution{Tsinghua University}
 \city{Beijing}
 \country{China}
 }
\email{fengtongtong@tsinghua.edu.cn}

\renewcommand{\shortauthors}{Trovato et al.}

\begin{abstract}
 
Long-Horizon (LH) tasks in Human-Scene Interaction (HSI) are complex multi-step tasks that require continuous planning, sequential decision-making, and extended execution across domains to achieve the final goal. However, existing methods heavily rely on skill chaining by concatenating pre-trained subtasks, with environment observations and self-state tightly coupled, lacking the ability to generalize to new combinations of environments and skills, failing to complete various LH tasks across domains. To solve this problem, this paper presents ALAS, a cross-domain learning framework for LH tasks via biologically inspired dual-stream disentanglement. Inspired by the brain's "where-what" dual pathway mechanism, ALAS comprises two core modules: i) an environment learning module for spatial understanding, which captures object functions, spatial relationships, and scene semantics, achieving cross-domain transfer through complete environment-self disentanglement; ii) a skill learning module for task execution, which processes self-state information including joint degrees of freedom and motor patterns, enabling cross-skill transfer through independent motor pattern encoding. We conducted extensive experiments on various LH tasks in HSI scenes. Compared with existing methods, ALAS can achieve an average subtasks success rate improvement of 23\% and average execution efficiency improvement of 29\%. 
\end{abstract}




\begin{teaserfigure}
  \includegraphics[width=1\textwidth]{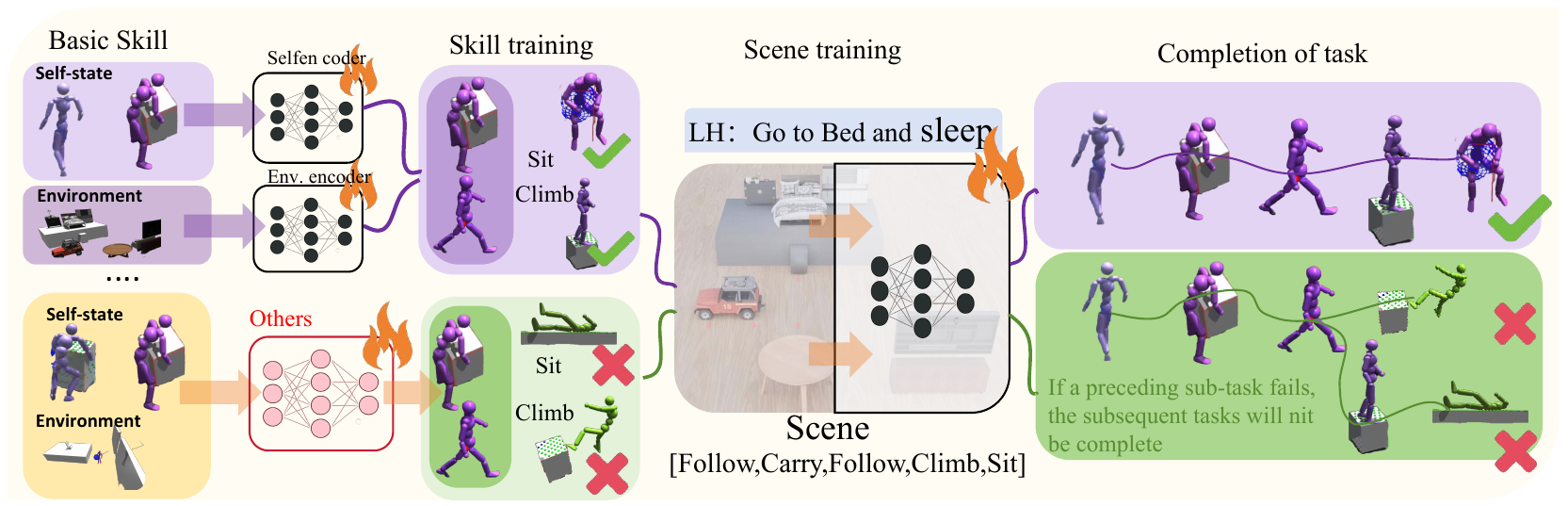}
  \caption{\small ALAS achieves generative generalization by learning fundamental subtasks in a single environment, enabling it to generalize to novel environments and accomplish Long-Horizon tasks that involve previously unseen subtasks.}
  \Description{Enjoying the baseball game from the third-base
  seats. Ichiro Suzuki preparing to bat.}
  \label{fig:teaser}
\end{teaserfigure}

\maketitle

\section{Introduction}
Long-Horizon (LH) tasks in Human-Scene Interaction (HSI) require continuous planning and cross-domain execution, posing challenges due to their complexity and need for environmental adaptation. These tasks have broad applications in robotics~\cite{qiu2024learning}, medical intervention~\cite{kim2024openvla}, and smart homes~\cite{kim2024openvla}, with canonical examples including dexterous hand manipulation~\cite{zhang2024contactdexnet} and humanoid whole-body control~\cite{sferrazza2024humanoidbench}. However, recent benchmarks show that HSI methods achieve low success rates on cross-domain tasks and demand extensive retraining~\cite{zhang2025interactanything,wang2024embodiedscan,xu2024interdreamer}, severely limiting real-world deployment.

Recent large-scale vision-language-action (VLA) models~\cite{black2024pi0visionlanguageactionflowmodel,team2025gemini} and agent-based manipulation~\cite{ni2024don} 
achieve strong results on long-horizon embodied tasks. 
However, both paradigms typically adopt monolithic or tightly coupled end-to-end designs, where perception and control remain entangled, thereby limiting cross-domain generalization and modular skill reuse.

To bridge these gaps, current approaches \cite{pan2025tokenhsi,li2024optimus,park2024iclr} focus on processing self-state information in unified representation spaces, while other solutions\cite{zhang2025interactanything,xiao2023unified,xu2024interdreamer} further encode self-state information mixed with environmental information. 
The efficacy of the \textit{decompose-reuse-compose} paradigm has been confirmed by various studies \cite{huang2020one,lan2023contrastive,xu2023composite,hu2024disentangled}, which also introduced a new modular learning paradigm for rapid adaptation to new skills by utilizing skill modules that have already been learned. 
In particular, CML~\cite{lan2023contrastive} and TokenHSI~\cite{pan2025tokenhsi} have explicitly demonstrated that such modular 
decomposition significantly outperforms standard end-to-end 
approaches in multi-task reinforcement learning (RL) and long-horizon 
HSI, respectively. 

Despite their promising performance, these methods suffer from the same architectural flaw: they adopt unified feature representation spaces that tightly couple environmental understanding with self-states. This flaw poses significant challenges in two main aspects: (1) Limited environmental transfer capability: When environmental changes occur (such as shifts from bright laboratory to dim factory settings), these systems cannot effectively separate the effects of environmental changes from self-state changes. This limitation necessitates relearning the entire perception-action mapping \cite{li2025controlling}, significantly constraining their cross-domain generalization capability. (2) Inefficient skill transfer capability: Current methods fail to achieve functional separation between perception and motor control. When encountering novel skills, even those involving similar motor patterns (such as grasping different objects), the system must retrain the entire perception-action network. This limitation makes it difficult to reuse, prevents effective reuse of learned motor skills, resulting in extremely low knowledge transfer efficiency due to a high risk of skill forgetting \cite{van2024continual}. Even
advanced modular approaches such as CML~\cite{lan2023contrastive} 
and TokenHSI~\cite{pan2025tokenhsi} still rely partly on unified feature spaces, thereby inheriting some of these limitations.

To address these challenges, this paper introduces \textbf{ALAS}: a biologically inspired functional disentanglement architecture that draws from the dorsal-ventral stream hypothesis in neuroscience \cite{ungerleider1982two}. According to this hypothesis, the brain's ventral \textit{what} pathway specializes in object recognition, while the dorsal \textit{where-how} pathway handles spatial processing and motor control. Unlike existing dual-stream approaches \cite{ibrayev2024toward} that separate visual modalities, ALAS introduces a functional disentanglement: the \textbf{Environmental Encoder} learns scene-invariant spatial relationships \cite{arkhangelsky2024causal} while the \textbf{Self-Encoder} captures body-schema-specific motor primitives.

Proposed method is extensively evaluated on various self-designed LH-embodied AI tasks, including cross-scene adaptation, novel skill adaptation, and particularly LH control tasks in complex environments. The contributions of this paper can be summarized as follows.

\begin{itemize}
\item Proposing the \textbf{ALAS disentangled architecture}, the first Embodied AI control framework in HSI based on biologically inspired cognitive principles. This architecture separates traditional unified encoding into specialized parallel processing of environmental perception streams and self-state perception streams.

\item Designing \textbf{specialized dual-stream encoders}, where the environmental encoder enhances \textbf{cross-domain transfer capability}, and the self-encoder achieves \textbf{cross-task skill reuse}. Both encoders are independently optimized and flexibly combined.

\item Establishing comprehensive benchmark scenarios for LH tasks through designed progressive LH task benchmarks, and validating the effectiveness of ALAS on these benchmarks. Compared to existing methods, ALAS achieves a \textbf{$2\times$ improvement in cross-domain adaptation capability} and a \textbf{$1.5\times$ improvement in skill reuse efficiency}.
\end{itemize}

\section{Related Works}
\subsection{Human-Scene Interaction}

HSI focuses on enabling embodied agents to interact naturally and effectively with complex 3D environments. Existing approaches include unified representation learning (e.g., Chain of Contacts \cite{xiao2023unified}), which integrates contact and object encoding with LLM-based planning but exhibits limited generalizability due to tight coupling between perception and action components; staged processing (e.g., Dynamic HSI~\cite{jiang2024scaling}), which uses autoregressive diffusion for disentangled scene understanding and action generation, ensuring temporal coherence but at high computational cost; and end-to-end methods (e.g., TokenHSI~\cite{pan2025tokenhsi}, and \cite{zhang2025interactanything,xu2024interdreamer}), which synthesize motion from text using pre-trained models and object sensors but are limited to simple skill composition scenarios. A key limitation shared by these approaches is their tight coupling between perception and control modules, which hinders cross-domain transfer and skill reusability.

\begin{figure*}[h]  
  \centering
  \includegraphics[width=\textwidth]{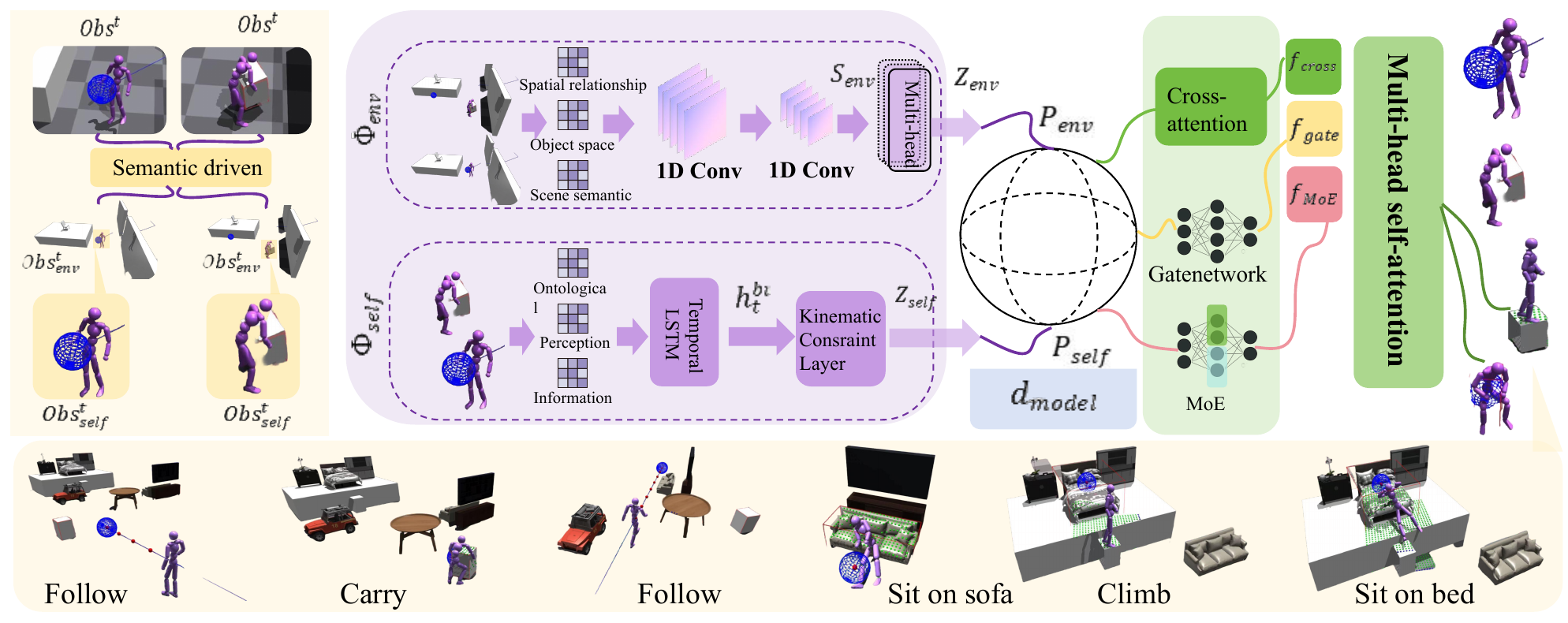}  
  \caption{Illustrating the operational workflow of the ALAS, Raw observation $\text{obs}^t$ is semantically disentangled into environmental $\text{obs}^t_{\text{env}}$ and self-state $\text{obs}^t_{\text{self}}$ components. Environmental encoder $\Phi_{\text{env}}$ and self-encoder $\Phi_{\text{self}}$ process respective inputs, with projection layers $P_{\text{env}}$ and $P_{\text{self}}$ mapping outputs to unified $d_{\text{model}}$ space. Multi-strategy adaptive fusion integrates features via three components: MoE fusion, gated fusion network, and cross-attention fusion module, producing outputs $(f_{\text{MoE}}, f_{\text{gate}}, f_{\text{cross}})$. These fused representations undergo Transformer multi-head self-attention before feeding into policy and value heads. }  
  \label{2}  
\end{figure*}
\subsection{Long-Horizon Task}
LH tasks in HSI require agents to perform multi-step reasoning and manage long-term dependencies \cite{li2024optimus}. Current approaches include hierarchical planning (e.g., MLLM-based instruction parsing with visual encoders \cite{li2025optimus2,zheng2023steve}), which decomposes tasks into subgoals but suffers from low skill prediction accuracy; memory augmentation (e.g., hierarchical memory and knowledge graphs \cite{li2024optimus}), which models long-term dependencies yet lack dynamic adaptation; and causal modeling \cite{li2025optimus3}, which enhances policy learning through observation-action causality but requires high computational resources and relies on limited training data. These methods are limited by their reliance on static representations, which constrains cross-domain transfer, policy reuse, and adaptation to dynamic interaction scenarios.

\subsection{Disentangled Learning}

Disentangled representation learning addresses these limitations by decomposing complex systems into independent, interpretable modules, improving generalization and controllability \cite{ada2024diffusion}. Key approaches include mutual information-based disentanglement (e.g., \cite{hu2024disentangled}), which minimizes mutual information between skill components but requires domain-specific prior knowledge; factorized representation learning (e.g., $\beta$-VAE framework \cite{uppal2025denoising}), which uses disentanglement regularization; and variational disentanglement (e.g., \cite{bhowal2024variational}), which optimizes a variational lower bound. An alternative method \cite{yang2025task} employs Wasserstein distance for stable disentanglement, though it remains theoretical, while \cite{yang2025task} also identifies valuable factors at high computational cost. However, these methods focus on static factor separation, which are ill-suited for the dynamic, continuous interactions and generative adaptation required in LH embodied tasks.

 \section{Method}
 ALAS employs a dual-encoder design with environmental encoder $\Phi_{\text{env}}$ and self-encoder $\Phi_{\text{self}}$ to disentangle environmental perception from self-state representation. Their outputs are fused via a multi-strategy adaptive mechanism and processed by the shared transformer encoder $\phi$ to enhance perception-control collaboration.

\subsection{ Observation Space Reconstruction Model}

ALAS employs observation disentanglement, modeling unified observation space as a Dual-Stream Separation Process (DSP). The disentanglement objective minimizes mutual information between environmental and self-state representations, quantified as $D = \sum_{t=0}^{T} \gamma^t I(obs^t_{env}, obs^t_{self})$, implemented using correlation-based mutual information estimators.

\subsection{ Disentangled Dual-Encoder}
ALAS is a biologically inspired, disentangled dual-encoder architecture that separates the traditional unified encoding pathway into two specialized processing streams for environmental perception and self-state representation. Figure 2 illustrates the proposed disentanglement module, which consists of four key components:

\textit{Environmental encoder $\Phi_{env}$.} The environmental encoder processes spatial information such as object positions and scene semantics. Given environmental observations $obs^t_{env} \in \mathbb{R}^{T \times d_{env}}$, we adopt parallel convolutional layers for feature extraction. Feature extraction is performed as:
\begin{equation}
S_{env} = \text{Concat}[\text{Conv1D}_k(obs^t_{env})]
\end{equation}

Features are aggregated through multi-head self-attention for spatial feature aggregation:
\begin{equation}
\begin{split}
z_{env} &= \text{LayerNorm}(\text{MultiHeadAttn}(S_{env}, S_{env}, S_{env}) \\ &\quad + S_{env})
\end{split}
\end{equation}

The environmental encoder is paired with decoder $\text{Decoder}_{env}$ for reconstruction-based pre-training.

\textit{Self-encoder $\Phi_{self}$.} The self-encoder processes self-state information $obs^t_{self} \in \mathbb{R}^{T \times d_{self}}$ including joint angles and velocities. Since accurate self-state understanding requires bidirectional temporal context for accurate motion understanding, the self-encoder employs a recurrent neural architecture with bidirectional processing capabilities. The model is defined as:
\begin{equation}
h^{bi}_t = [h^f_t; h^b_t]
\end{equation}
where $h^f_t$ and $h^b_t$ represent forward and backward temporal representations, respectively.

The kinematic constraint layer ensures outputs remain within physically feasible ranges through a element-wise soft gating mechanism:
\begin{equation}
z_{self} = h^{bi}_t \odot \sigma(W_k h^{bi}_t + b_k)
\end{equation}
where $\sigma$ is the sigmoid function, and $W_k$ and $b_k$ are learnable parameters. The kinematic constraint layer ensures the physical feasibility of generated actions.

The self-encoder is paired with a temporal prediction network $f_{pred}$ for sequence prediction-based pre-training, which learns to predict future self-state representations from current ones.
\textit{Feature projection layers $P_{env}$ and $P_{self}$.} Two independent linear layers map the encoder outputs to a unified $d_{model}$ dimensional space:
\begin{equation}
f_{env} = P_{env}(z_{env}), \quad f_{self} = P_{self}(z_{self})
\end{equation}
where $f_{env}, f_{self} \in \mathbb{R}^{d_{model}}$ are the projected features used for fusion.

\subsection{Multi-Strategy Adaptive Fusion Mechanism}
To effectively integrate heterogeneous features from the environmental encoder and self-encoder, \textbf{ALAS} incorporates a multi-strategy adaptive fusion mechanism that combines three complementary fusion strategies. According to Figure 2, the adopted fusion mechanism comprises three core components:

\textit{Cross-attention fusion module.} This module \cite{vaswani2017attention} is selected for its capability to enable internal states to actively query key information from the environment, thereby achieving state-driven dynamic feature alignment. It uses self-state features $f_{self} \in \mathbb{R}^{d_{model}}$ as Query, and environment features $f_{env} \in \mathbb{R}^{d_{model}}$ as Key and Value, achieving dynamic weight allocation through a multi-head attention mechanism:

\begin{equation}
\begin{split}
f_{cross} &= \text{MultiHead}(f_{self}, f_{env}, f_{env}) \\
&= \text{Concat}(\text{head}_1, \ldots, \text{head}_h)W^O
\end{split}
\end{equation}
where each attention head:
\begin{equation}
\text{head}_i = \text{Attention}(f_{self} W^Q_i, f_{env} W^K_i, f_{env} W^V_i)
\end{equation}


\textit{Gated Fusion Network.} This module dynamically modulates contribution weights between environmental perception and self-state features to prevent imbalance, using learnable gating units. It is implemented via a multi-layer MLP \cite{tolstikhin2021mlp} with decreasing hidden units, matching the fused feature dimension. The gated fusion strategy is defined as:
\begin{multline}
f_{gate} = \sigma(W_g [f_{env}; f_{self}] + b_g) \odot f_{env} \\
+ (1 - \sigma(W_g [f_{env}; f_{self}] + b_g)) \odot f_{self}
\end{multline}

\textit{Mixture of Experts (MoE) fusion module .} This module is adopted for its capacity to dynamically select optimal fusion experts based on task characteristics and environmental complexity, enabling adaptive feature integration. It designs multiple specialized fusion experts \cite{zadouri2023pushing}, each modeled by a multi-layer MLP network with a hierarchical structure, dynamically selecting the most suitable expert for feature fusion through a routing network. The mixture of experts' fusion is represented as:
\begin{equation}
f_{moe} = \sum_{i=1}^{4} w_i \cdot E_i(f_{env}, f_{self})
\end{equation}
where the routing weights are
\begin{equation}
w_i = \text{Softmax}(W_r [f_{env}; f_{self}] + b_r)_i
\end{equation}

The three fusion strategies are combined through a learnable weighted combination:
\begin{equation}
f_{fused} = \alpha \cdot f_{cross} + \beta \cdot f_{gate} + \gamma \cdot f_{moe}
\end{equation}
where $\alpha, \beta, \gamma$ are learnable parameters that balance the contributions of different fusion strategies.

\begin{figure}[t]
    \centering
    \includegraphics[width=0.45\textwidth]{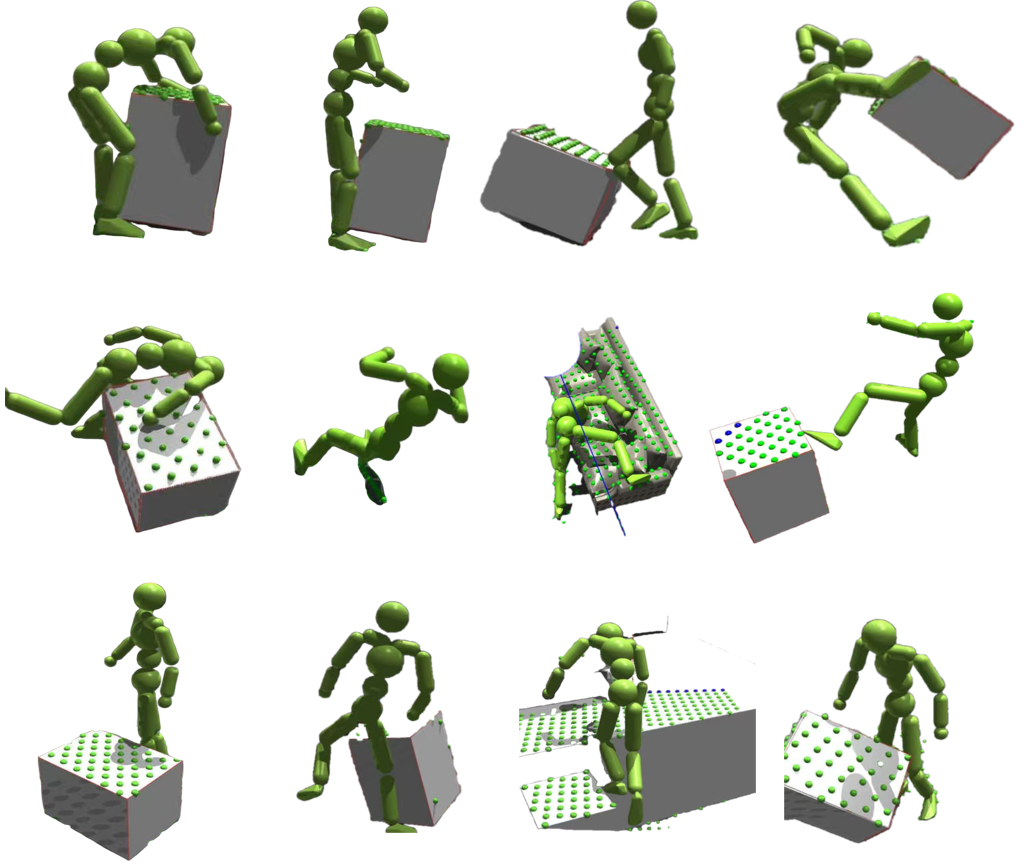}
    \caption{Previous methods lack the ability of cross-domain generalization, which leads to unsatisfactory performance on may unseen tasks and thus results in task failure. Existing issues include box overturning, unobstructed failing of the SMPL avatar, being pushed by boxes, and the inability to proceed with subsequent actions.}
    \label{fig:skill_comparison}
\end{figure}

\textit{Shared Transformer Encoder.} The fused features are processed by a shared transformer encoder $\phi$ to enhance perception-control collaboration:
\begin{equation}
h_{transformer} = \phi(f_{fused})
\end{equation}
where $\phi$ consists of multiple transformer layers with self-attention mechanisms to capture long-range dependencies and temporal relationships.

\textit{Policy and Value Heads.} The transformer output is fed into separate policy and value heads for action prediction and value estimation:
\begin{equation}
\begin{split}
\pi(a|s) &= \text{PolicyHead}(h_{transformer}) \\
V(s) &= \text{ValueHead}(h_{transformer})
\end{split}
\end{equation}

where $\pi(a|s)$ represents the action probability distribution and $V(s)$ represents the state value function.

\subsection{ Progressive Training Protocol}
To fully leverage the advantages of disentangled architecture and ensure the specialized characteristics of each module, ALAS designed a comprehensive progressive training protocol and specialized regularization mechanisms. The adopted training protocol involves three progressive stages, each with clear training objectives and parameter update strategies:

\textit{Independent Pre-training Stage.} In this stage, the environmental encoder $\Phi_{env}$ and self-encoder $\Phi_{self}$ are trained independently to establish their respective feature representation capabilities. The environmental encoder is pre-trained through the scene reconstruction loss:
\begin{equation}
\mathcal{L}_{env} = \|\text{Decoder}_{env}(\Phi_{env}(obs^t_{env})) - obs^t_{env}\|_2^2
\end{equation}

The self-encoder is pre-trained through action sequence prediction tasks:
\begin{equation}
\mathcal{L}_{self} = \sum_{t=1}^{T-1} \|\Phi_{self}(obs^{t+1}_{self}) - f_{pred}(\Phi_{self}(obs^t_{self}))\|_2^2
\end{equation}
where $f_{pred}$ is the temporal prediction network.This stage establishes domain-specific representation foundations.

\textit{Fusion Layer Optimization Stage.} In this stage, the pre-trained encoder parameters $\theta_{env}, \theta_{self}$ are frozen to preserve learned representations, focusing on training the feature fusion layer and Transformer encoder $\phi$:
\begin{equation}
\mathcal{L}_{fusion} = \mathcal{L}_{task} + \lambda_{quality} \mathcal{L}_{fusion\_quality}
\end{equation}
where $\mathcal{L}_{task}$ represents the standard reinforcement learning objective (e.g., policy gradient loss for PPO), which guides the agent to maximize expected cumulative rewards.

where the fusion quality loss is defined as:
\begin{equation}
\begin{split}
\mathcal{L}_{fusion\_quality} &= \|f_{cross} - (f_{env} + f_{self})\|_2^2 \\
&\quad + \lambda_{disentangle} \cdot I(z_{env}, z_{self})
\end{split}
\end{equation}
where $I(z_{env}, z_{self})$ represents the mutual information between environmental and self-state features. The first term ensures fusion consistency, while the second term maintains disentanglement by minimizing mutual information between representations.

\textit{End-to-End Joint Optimization Stage.} In the end-to-end joint optimization stage, all network parameters are unfrozen for end-to-end joint optimization, while introducing specialized preservation regularization:
\begin{equation}
\begin{split}
\mathcal{L}_{total} &= \mathcal{L}_{task} + \lambda_{disentangle} \cdot I(z_{env}, z_{self}) \\
&\quad + \sum_{i} \lambda_i \mathcal{R}_i
\end{split}
\end{equation}
where the regularization terms include:
\begin{align}
\mathcal{R}_1 &= \|\theta_{env} - \theta_{env}^*\|_2^2 \quad \text{(encoder preservation)} \\
\mathcal{R}_2 &= \|\theta_{self} - \theta_{self}^*\|_2^2 \quad \text{(encoder preservation)} \\
\mathcal{R}_3 &= \sum_{i \neq j} \|f_i - f_j\|_2^2 \quad \text{(fusion diversity)}
\end{align}
where $f_i, f_j \in \{f_{cross}, f_{gate}, f_{moe}\}$ and $\mathcal{R}_3$ encourages different fusion strategies to learn complementary representations. The regularization weights $\lambda_i$ (where $i \in \{1,2,3\}$) are hyperparameters that balance the contributions of different regularization terms. $\theta_{env}^*$ and $\theta_{self}^*$ are the pre-trained encoder parameters, and the disentanglement regularization term $I(z_{env}, z_{self})$ ensures continued minimization of mutual information between environmental and self-state representations throughout the training process.

\section{Experiment}
We conduct comprehensive experiments to evaluate our method across foundational skill learning and Long-Horizon (LH) task execution. To ensure the statistical significance of our results and account for the inherent stochasticity in reinforcement learning, we evaluated ALAS and all baseline methods across 10 independent random seeds. All performance metrics are reported as $\text{mean} \pm \text{standard deviation}$ to provide a rigorous assessment of model stability. This protocol ensures that the observed improvements are not the result of fortuitous weight initialization but represent the robust capability of our framework.

Our experiments are conducted entirely on three LH tasks that we designed:

\textbf{LH1: ``Sit on Chair!''} This LH task comprises a sequence of four fundamental skills: \textit{Follow, Carry, Climb,} and \textit{Sit}, where the target object for \textit{Sit} is a Chair.

\textbf{LH2: ``Sit on Sofa!''} This LH task similarly comprises a sequence of four fundamental skills: \textit{Follow, Carry, Follow,} and \textit{Sit}, where the target object for \textit{Sit} is a Sofa.

\textbf{LH3: ``Go to Bed!''} This LH task comprises a sequence of five fundamental skills: \textit{Follow, Carry, Follow, Climb,} and \textit{Sit}, where the target object for \textit{Sit} is a Bed.

Our object assets are sourced from the 3D-FRONT dataset \cite{fu20213d}, while the motion data is inherited from TokenHSI~\cite{pan2025tokenhsi}.

\begin{figure}[t]
    \centering
    \includegraphics[width=0.45\textwidth]{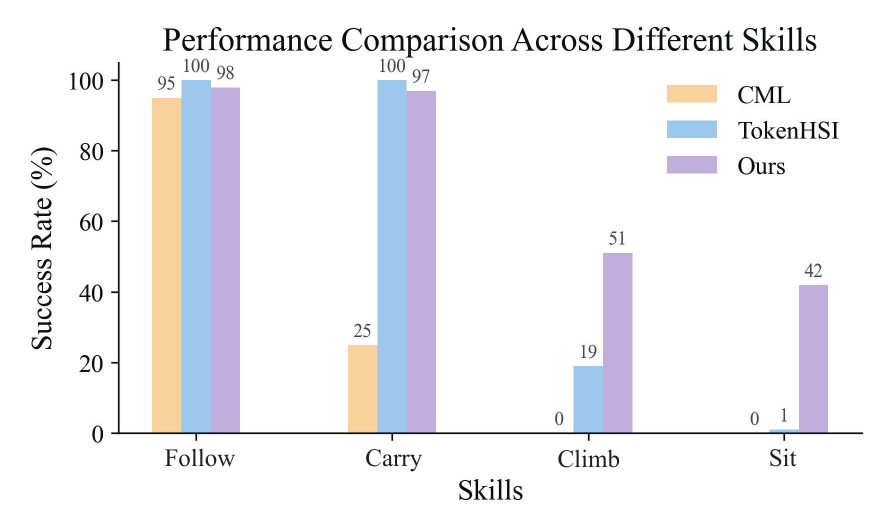}
    \caption{Success rate comparison across different skills among CML, TokenHSI, and our ALAS framework.}
    \label{fig:skill_comparison}
\end{figure}

\begin{table}[t]
\centering
\renewcommand{\arraystretch}{1.3} 
\caption{Success rates (Mean $\pm$ Std) for foundational skills and composite task completion. Results are evaluated across 10 independent random seeds (100 trials in total).}
\begin{tabular}{lccc}
\hline
\textbf{Task / Skill} & \textbf{CML}~\cite{xu2023composite} & \textbf{TokenHSI}~\cite{pan2025tokenhsi} & \textbf{Ours (ALAS)} \\
\hline
Follow & $0.95 \pm 0.03$ & $1.00 \pm 0.01$ & $0.98 \pm 0.02$ \\
Carry  & $0.25 \pm 0.08$ & $1.00 \pm 0.02$ & $0.97 \pm 0.03$ \\
Climb  & $0.00 \pm 0.00$ & $0.19 \pm 0.06$ & \cellcolor{gray!15}$\mathbf{0.51 \pm 0.04}$ \\
Sit    & $0.00 \pm 0.00$ & $0.01 \pm 0.02$ & \cellcolor{gray!15}$\mathbf{0.42 \pm 0.05}$ \\
\hline
\textbf{LH1} & $0.30 \pm 0.07$ & $0.55 \pm 0.09$ & \cellcolor{gray!15}$\mathbf{0.72 \pm 0.04}$ \\
\hline
\end{tabular}
\label{tab:success_rates}
\end{table}

\begin{table*}[htbp]
\centering
\renewcommand{\arraystretch}{1.2} 
\caption{Comparison of generalization performance among ALAS and baselines on LH tasks.}
\label{tab:experiment_results}
\resizebox{1.0\textwidth}{!}{%
\begin{tabular}{c|l|ccccccccc}
\hline
\textbf{Exp.} & \textbf{Method} & \textbf{Follow} & \textbf{Carry} & \textbf{Follow} & \textbf{Climb} & \textbf{Sit} & \textbf{Time(s)} & \textbf{LH Success} & \textbf{SGR} & \textbf{EGR} \\
\hline
\multirow{4}{*}{LH2} 
 & HLR & $0.91 \pm 0.05$ & $0.42 \pm 0.09$ & $0.10 \pm 0.03$ & {-} & $0.00 \pm 0.00$ & $115.2 \pm 12.4$ & $0.28 \pm 0.08$ & $0.00 \pm 0.00$ & $0.62 \pm 0.10$ \\
 & PULSE  & $0.88 \pm 0.06$ & $0.35 \pm 0.11$ & $0.08 \pm 0.04$ & {-} & $0.00 \pm 0.00$ & $128.5 \pm 15.1$ & $0.21 \pm 0.07$ & $0.00 \pm 0.00$ & $0.55 \pm 0.12$ \\
 & TokenHSI & $1.00 \pm 0.01$ & $0.56 \pm 0.08$ & $0.13 \pm 0.03$ & {-} & $0.01 \pm 0.02$ & $99.0 \pm 8.5$ & $0.42 \pm 0.09$ & $0.01 \pm 0.02$ & $0.76 \pm 0.07$ \\
 & \cellcolor{gray!20} \textbf{Ours} & \cellcolor{gray!20} \textbf{1.00 $\pm$ 0.01} & \cellcolor{gray!20} \textbf{0.96 $\pm$ 0.03} & \cellcolor{gray!20} \textbf{0.67 $\pm$ 0.05} & \cellcolor{gray!20} {-} & \cellcolor{gray!20} \textbf{0.16 $\pm$ 0.05} & \cellcolor{gray!20} \textbf{85.0 $\pm$ 6.2} & \cellcolor{gray!20} \textbf{0.70 $\pm$ 0.05} & \cellcolor{gray!20} \textbf{0.08 $\pm$ 0.02} & \cellcolor{gray!20} \textbf{0.97 $\pm$ 0.02} \\
\hline
\multirow{4}{*}{LH3} 
 & HLR & $0.89 \pm 0.06$ & $0.38 \pm 0.10$ & $0.15 \pm 0.04$ & $0.05 \pm 0.03$ & $0.00 \pm 0.00$ & $120.8 \pm 14.2$ & $0.25 \pm 0.07$ & $0.45 \pm 0.08$ & $0.58 \pm 0.11$ \\
 & PULSE  & $0.85 \pm 0.08$ & $0.31 \pm 0.09$ & $0.12 \pm 0.05$ & $0.02 \pm 0.02$ & $0.00 \pm 0.00$ & $135.2 \pm 18.3$ & $0.18 \pm 0.08$ & $0.38 \pm 0.09$ & $0.50 \pm 0.13$ \\
 & TokenHSI & $1.00 \pm 0.01$ & $0.50 \pm 0.07$ & $0.21 \pm 0.03$ & $0.20 \pm 0.06$ & $0.00 \pm 0.00$ & $102.9 \pm 9.2$ & $0.38 \pm 0.09$ & $0.67 \pm 0.07$ & $0.69 \pm 0.08$ \\
 & \cellcolor{gray!20} \textbf{Ours} & \cellcolor{gray!20} \textbf{1.00 $\pm$ 0.01} & \cellcolor{gray!20} \textbf{0.95 $\pm$ 0.03} & \cellcolor{gray!20} \textbf{0.50 $\pm$ 0.06} & \cellcolor{gray!20} \textbf{0.40 $\pm$ 0.06} & \cellcolor{gray!20} \textbf{0.10 $\pm$ 0.04} & \cellcolor{gray!20} \textbf{97.6 $\pm$ 7.1} & \cellcolor{gray!20} \textbf{0.59 $\pm$ 0.06} & \cellcolor{gray!20} \textbf{0.13 $\pm$ 0.03} & \cellcolor{gray!20} \textbf{0.81 $\pm$ 0.05} \\
\hline
\end{tabular}
}
\end{table*}

\subsection{ Evaluation on Foundational Skill Learning and Task Completion}
\textbf{Experimental Setup.}  
To evaluate the robustness and universality of our disentangled architecture, we employ a progressive learning protocol where foundational skills \textit{Follow} and \textit{Carry} are established through comprehensive training, while \textit{Climb} and \textit{Sit} skills are acquired through compositional learning. This approach enables systematic assessment of skill generalization capabilities and adaptation to diverse environments.
The training procedure uses large-scale parallelization in 4,096 environments, employing PPO \cite{schulman2017proximal} with 10k iterative updates. We conducted 100 independent experimental trials to ensure statistical reliability, quantifying robustness and universality through the success rate means of all skills and L1 tasks. This rigorous evaluation framework provides a comprehensive assessment of the architecture's performance through systematic skill composition and environmental adaptation.

\begin{figure}[t]
    \centering
    \includegraphics[width=0.4\textwidth]{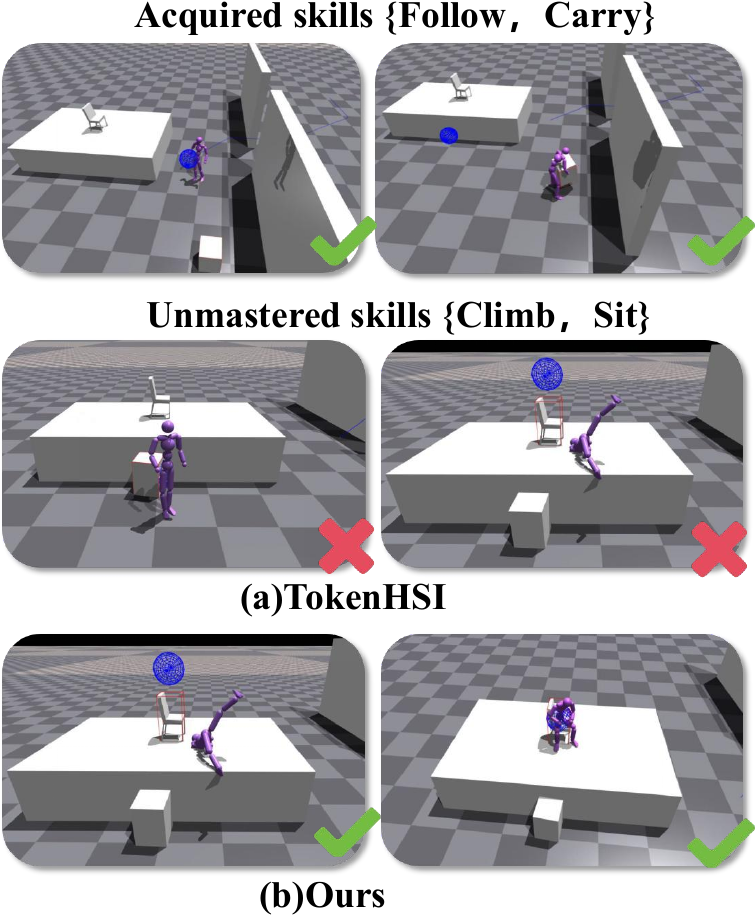}
    \caption{Skill acquisition performance comparison between ALAS and TokenHSI.} 
    \label{fig:skill_acquisition}
\end{figure}

\textbf{Baselines.} 
We train TokenHSI from scratch using our custom dataset. TokenHSI is a state-of-the-art full-body humanoid controller that learns a set of foundational skills comparable to ours. We also include CML~\cite{xu2023composite}, a composite motion learning baseline commonly used alongside TokenHSI, as an additional point of reference.

\textbf{Follow and Carry.} The success rate of \textit{Follow} is defined as maintaining the pelvis within a 30cm distance threshold from the target path in the XY plane. For the \textit{Carry} task, which can be decomposed into 'grasp' and 'transport' components, achieving only the grasp phase without successful transport to the designated target location is considered 0.5 task completion. \textit{Follow} task training utilized procedurally generated trajectories, while \textit{Carry} task training employed 9 boxes of varying dimensions. Subsequently, we trained on the compositional LH1 task combining these two primitives and evaluated performance on identical task compositions.

\textbf{Climb and Sit.} The success of \textit{Climb} is defined as reaching the target object with the pelvis positioned at or above the target elevation. Success of \textit{Sit} requires the pelvis to be positioned on the upper surface of the target object.

\textbf{LH1 task.} The Success rate for LH1 is the success rate of the sub-skill sequence. Due to the sequential nature of LH tasks, where skills must be executed in order, failure in a preceding task prevents the execution of subsequent tasks. Therefore, the skill sequence success rate serves as an excellent metric for evaluating the success rate of LH tasks.

\textbf{Results.}
Table~\ref{tab:success_rates} presents a comprehensive quantitative analysis of CML, TokenHSI, and our proposed ALAS. The results, averaged over 10 independent random seeds, underscore the superior generalization and stability of our approach. While all methods achieve comparable proficiency in basic pre-trained skills such as \textit{Follow} and \textit{Carry}, ALAS significantly outperforms the baselines as task complexity and domain shifts increase. Specifically, ALAS achieves success rates of $\mathbf{0.51 \pm 0.04}$ and $\mathbf{0.42 \pm 0.05}$ on the challenging \textit{Climb} and \textit{Sit} tasks, respectively, whereas CML and TokenHSI largely fail to generalize, with success rates approaching zero. Notably, for the overall Long-Horizon task (\textbf{LH1}), ALAS attains a completion rate of $\mathbf{0.72 \pm 0.04}$, surpassing CML by 42\% and TokenHSI by 17\%. Furthermore, the consistently lower standard deviations of ALAS across all metrics demonstrate its exceptional robustness against stochastic variations during training and execution, confirming that its performance gains are statistically significant rather than the result of fortuitous initialization.

\subsection{Long-Horizon Task Completion}

This section evaluates the ALAS framework's performance on Long-Horizon (LH) tasks, designed to test generalization across skills and environments. We focus on \textbf{skill generalization} and \textbf{environment generalization}, using LH2 and LH3 for assessment, which target adaptation to novel environments and task compositions. Generalization is evaluated over 100 test runs per task, measuring subtask success rates in diverse, unseen scenes to assess robustness.

\textbf{Task Execution Times.} Task execution time refers to the duration from the start of the current LH task to the initiation of the next LH task. The criteria for determining the execution of the next task include the occurrence of errors (such as falling) or exceeding the threshold time for task execution.

\textbf{Experiment setup.}
As described in Section 4.1, to validate the environment generalization capability of our framework, we employ the same progressive learning protocol on the LH1 task and directly evaluate its performance on the LH2 and LH3 tasks. This zero-shot evaluation on novel scene layouts allows for a more intuitive assessment of the environmental adaptability inherent in the ALAS framework. To further benchmark our method, we incorporate three representative baselines: \textbf{TokenHSI}~\cite{pan2025tokenhsi}, which serves as a modular high-level interaction baseline, and two Hierarchical Reinforcement Learning (HRL) methods, \textbf{PULSE}~\cite{lu2024pulse} and \textbf{Hybrid Latent Representation(HLR)}~\cite{bae2025versatile}. 

While HRL methods excel at temporal abstraction through goal-conditioned structures, they typically operate on a coupled observation space where environmental geometry and agent states are entangled. Similarly, although TokenHSI facilitates modular interaction, its generalization to unseen tasks remains limited by the lack of explicit functional disentanglement. Since we establish foundational skills such as \textit{Follow} and \textit{Carry} through comprehensive pre-training, we evaluate skill generalization and transferability by observing completion rates on the previously unseen sub-tasks \textit{Climb} and \textit{Sit}. As demonstrated in the following results, ALAS maintains superior performance across these novel tasks, whereas the three baselines struggle to adapt their learned policies to new spatial configurations.

\begin{figure}[h]  
  \centering
  \includegraphics[width=0.4\textwidth]{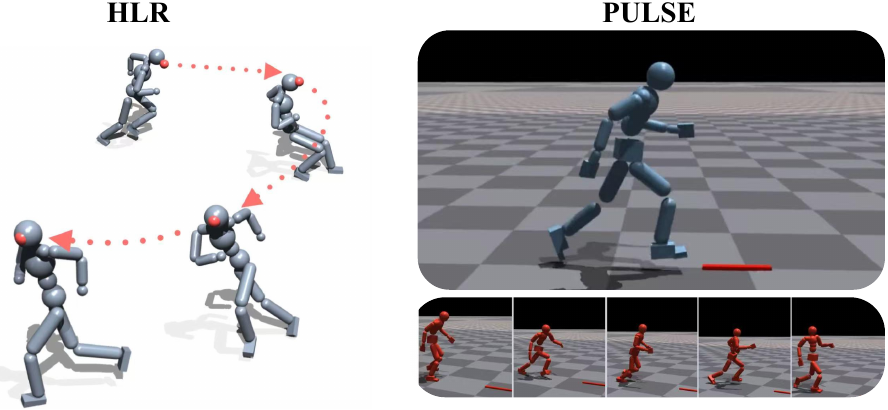}  
  \caption{Within the HRL framework, \textbf{PULSE} constructs a continuous latent space as the high-level action via distillation from a teacher model to achieve universal motion representation, whereas \textbf{HLR} builds a hybrid continuous-discrete latent space as the high-level action to balance stability and smoothness.}  
  \label{2}  
\end{figure}

\textbf{Generalization Rate Definition.} Based on our experimental data, we formally define the Environment Generalization Rate (EGR) and Skill Generalization Rate (SGR) as follows:
\begin{equation}
EGR = \frac{S_{Li}}{S_{L1}}, i \in {2,3}
\label{eq:egr}
\end{equation}

\begin{equation}
SGR = \frac{(S_{climb} + S_{sit})/2}{(S_{follow} + S_{carry})/2}
\label{eq:sgr}
\end{equation}
where $S_{Li}, i \in \{1,2,3\}$ represents the success rate of LH tasks, and the testing on LH2 and LH3 involves direct transfer from the LH1 environment training, demonstrating its rationality. Similarly, $S_{climb}$ etc. represent the success rates of skills, where \textit{Climb} and \textit{Sit} are composed from foundational \textit{Follow} and \textit{Carry} skills; therefore, we define the skill generalization rate using this formula.

\begin{figure*}[t]  
  \centering
  \includegraphics[width=0.90\textwidth]{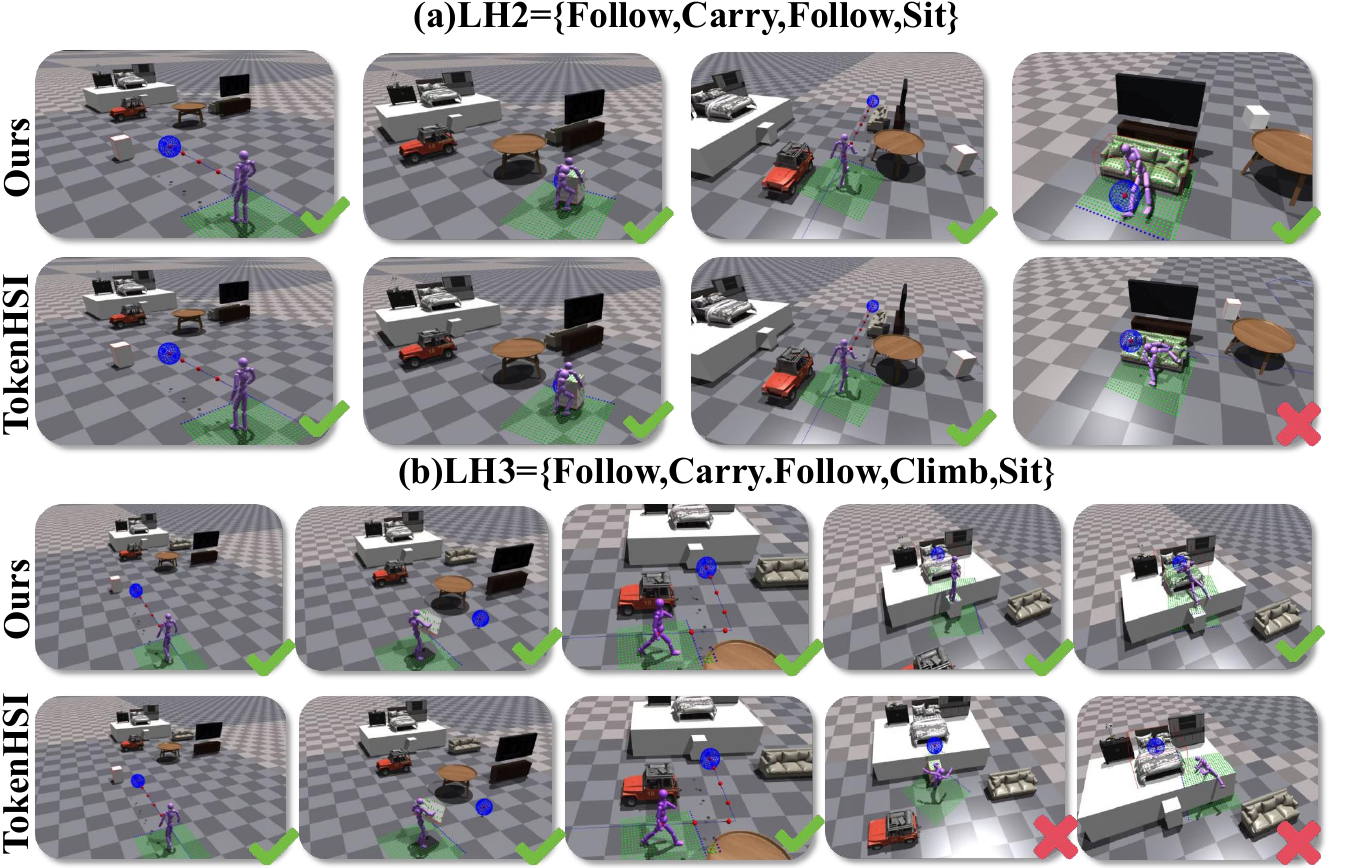}  
  \caption{Generalization comparison between ALAS and TokenHSI on LH tasks, where (a) and (b) represent tasks composed of sequences of four and five foundational skills, respectively. We only pre-trained the first two actions, \textit{Follow} and \textit{carry} on LH1 tasks, and tested skill generalization and environmental generalization in new scenarios.}  
  \label{2}  
\end{figure*}

\textbf{Results.}
Figure~\ref{2} visually underscores the superior skill composition of ALAS compared to baseline methods. A detailed quantitative comparison, including sub-task success rates, overall LH task completion, execution efficiency, and generalization metrics, is provided in Table~\ref{tab:experiment_results}. 

The results demonstrate that ALAS consistently outperforms all baselines, particularly in terms of operational efficiency and robustness. Specifically, ALAS reduces the average execution time by approximately 14s (LH2) and 5s (LH3) compared to the strongest baseline, TokenHSI. This efficiency gain is crucial, as longer execution times in complex HSI scenarios often lead to task failure due to the inherent 120s timeout constraint. In contrast, hierarchical methods such as HLR and PULSE exhibit significantly higher latency, frequently exceeding the time limit.

In terms of mission success, ALAS improves the LH task success rate by 28\% in LH2 and 21\% in LH3 over TokenHSI, effectively balancing swiftness with reliability. Notably, our framework achieves exceptional environment generalization rates (EGR) of 0.97 (LH2) and 0.81 (LH3), alongside skill generalization rates (SGR) of 0.08 (LH2) and 0.13 (LH3). These metrics significantly exceed those of HLR, PULSE, and TokenHSI, which struggle to adapt to novel spatial configurations. These results collectively validate that ALAS's functional disentanglement and adaptive fusion mechanisms provide superior composition capabilities for long-horizon HSI tasks.

\begin{table}[t] 
\small 
\setlength{\tabcolsep}{4pt} 
\renewcommand{\arraystretch}{1.2}
\centering
\caption{Ablation study on LH3 task. A1 and A2 verify the necessity of dual-stream disentanglement, while A3 evaluates the strategy fusion mechanism.}
\resizebox{0.48\textwidth}{!}{%
\begin{tabular}{c|l|ccc}
\hline
\textbf{ID} & \textbf{Configuration} & \textbf{EGR.} & \textbf{SGR.} & \textbf{LH.} \\
\hline
\textbf{Full} & \textbf{ALAS (Full Framework)} & \textbf{0.81} & \textbf{0.13} & \textbf{0.58} \\
\hline
A1 & w/o Env. Branch ($\Phi_{\text{env}}$) & \cellcolor{gray!15}0.64 & 0.12 & 0.41 \\
A2 & w/o Skill Branch ($\Phi_{\text{self}}$) & 0.74 & \cellcolor{gray!15}0.05 & 0.38 \\
A3 & w/o Adaptive Fusion (Concat) & 0.78 & 0.09 & 0.48 \\
\hline
\end{tabular}
}
\label{tab:ablation}
\end{table}

\subsection{ Ablation Experiment}

To evaluate the individual contributions of key modules in our ALAS framework, we perform a comprehensive ablation study. Each variant is constructed by disabling or removing a specific component from the full model while keeping all other settings fixed. The experiments are conducted on  LH3 tasks composed of foundational skill primitives (e.g., \textit{follow}, \textit{carry}, \textit{climb}, \textit{sit}) in diverse environments.

\textbf{Ablation Study Setup.} To evaluate the individual contribution of each component within the ALAS framework and justify the necessity of functional disentanglement, we conducted a series of ablation experiments on the most challenging \textit{LH3} task. Following the progressive learning protocol in Section 4.1, we recorded three key metrics: \textbf{Environment Generalization Rate (EGR)}, \textbf{Skill Generalization Rate (SGR)}, and the \textbf{Overall LH Success Rate}. Specifically, we compared our full framework against the following variants:
\begin{itemize}
    \item \textbf{A1 (w/o Env. Branch)}: Disables the environmental encoder $\Phi_{\text{env}}$ to assess the impact of scene-geometry disentanglement on spatial adaptability.
    \item \textbf{A2 (w/o Skill Branch)}: Disables the self-state encoder $\Phi_{\text{self}}$ to examine whether agent-specific motor patterns can be transferred without an independent latent stream.
    \item \textbf{A3 (w/o Adaptive Fusion)}: Replaces the gated MoE fusion mechanism with a simple concatenation layer to verify the necessity of dynamic policy coordination.
\end{itemize}
The quantitative results, averaged over 10 independent trials, are summarized in Table~\ref{tab:ablation}.

\textbf{Results.}
As illustrated in Table~\ref{tab:ablation}, the ablation results underscore the functional significance of each component within the ALAS framework. Disabling the environmental encoder (\textbf{A1}) causes the Environment Generalization Rate (EGR) to drop sharply from 0.81 to 0.64, signifying that specialized perception of scene geometry is vital for spatial adaptability. Similarly, removing the self-encoder (\textbf{A2}) leads to a substantial degradation in the Skill Generalization Rate (SGR), plummeting from 0.13 to 0.05, which confirms that isolating agent-specific motor patterns is essential for zero-shot skill transfer. Furthermore, replacing our adaptive fusion mechanism with a simple concatenation layer (\textbf{A3}) results in a noticeable decline in the overall LH success rate (from 0.58 to 0.48). This gap highlights that the gated coordination between streams is necessary for managing complex transitions in long-horizon sequences. Collectively, these findings validate that our dual-stream disentanglement and adaptive fusion are non-redundant and indispensable for robust HSI task performance.

\section{Conclusion and Future Work}
In this work, we presented \textbf{ALAS}, a biologically inspired dual-stream disentanglement framework that explicitly separates environment understanding from self-state encoding. This design enables \textbf{cross-domain transfer, modular skill reuse, and efficient long-horizon task composition}.
Extensive experiments on diverse HSI scenarios demonstrate that ALAS achieves substantial improvements of 23\% in subtask success rate and 29\% in execution efficiency, along with stronger generalization over state-of-the-art modular baselines.
While our current implementation relies on a pre-defined skill set, future work will explore open-ended skill discovery from unlabeled data and real-world deployment under dynamic environments. We believe ALAS provides a promising step toward scalable, generalizable embodied intelligence in complex human-scene interactions.

\bibliographystyle{ACM-Reference-Format}
\bibliography{sample-base}










\end{document}